\documentclass{article}

\usepackage{arxiv}

\usepackage[utf8]{inputenc} 
\usepackage[T1]{fontenc}    
\usepackage{hyperref}       
\usepackage{url}            
\usepackage{booktabs}       
\usepackage{amsfonts}       
\usepackage{nicefrac}       
\usepackage{microtype}      
\usepackage{lipsum}
\usepackage{graphicx}
\usepackage{subfig}
\usepackage{lineno}
\usepackage{tabularx}
\graphicspath{ {./images/} }

\usepackage{algorithm}
\usepackage{algpseudocode}

\title{So2Sat POP - A Curated Benchmark Data Set for Population Estimation from Space on a Continental Scale}

\author{
  Sugandha Doda \\
  Chair of Data Science in Earth Observation, \\
  Department of Aerospace and Geodesy, \\
  Technical University of Munich, \\
  Arcisstra{\ss}e 21, Munich,80333, Germany \\
   \And
  Yuanyuan Wang \\
  Chair of Data Science in Earth Observation, \\
  Department of Aerospace and Geodesy, \\
  Technical University of Munich, \\
  Arcisstra{\ss}e 21, Munich,80333, Germany \\
  \And
  Matthias Kahl \\
  Chair of Data Science in Earth Observation, \\
  Department of Aerospace and Geodesy, \\
  Technical University of Munich, \\
  Arcisstra{\ss}e 21, Munich,80333, Germany \\
  \And
  Eike Jens Hoffmann \\
  Chair of Data Science in Earth Observation, \\
  Department of Aerospace and Geodesy, \\
  Technical University of Munich, \\
  Arcisstra{\ss}e 21, Munich,80333, Germany \\
  \And
  Kim Ouan \\
  Chair of Data Science in Earth Observation, \\
  Department of Aerospace and Geodesy, \\
  Technical University of Munich, \\
  Arcisstra{\ss}e 21, Munich,80333, Germany \\
  \And
  Hannes Taubenböck \\
  German Remote Sensing Data Center, \\
  German Aerospace Center,  \\
  M\"unchener Stra{\ss}e 20, We{\ss}ling, 82234, Germany \\
  Institute for Geography and Geology, \\
  Julius-Maximilians-Universit\"t W\"urzburg, \\
  W\"urzburg 97074, Germany
  \And
  Xiao Xiang Zhu\thanks{corresponding author: Xiao Xiang Zhu (xiaoxiang.zhu@dlr.de)}\\
  Chair of Data Science in Earth Observation, \\
  Department of Aerospace and Geodesy, \\
  Technical University of Munich, \\
  Arcisstra{\ss}e 21, Munich,80333, Germany \\
}

\begin{document}
\maketitle
\begin{abstract}
Obtaining a dynamic population distribution is key to many decision-making processes such as urban planning, disaster management and most importantly helping the government to better allocate socio-technical supply. For the aspiration of these objectives, good population data is essential. The traditional method of collecting population data through the census is expensive and tedious. In recent years, statistical and machine learning methods have been developed to estimate population distribution. Most of the methods use data sets that are either developed on a small scale or not publicly available yet. Thus, the development and evaluation of new methods become challenging. We fill this gap by providing a comprehensive data set for population estimation in 98 European cities. The data set comprises a digital elevation model, local climate zone, land use proportions, nighttime lights in combination with multi-spectral Sentinel-2 imagery, and data from the Open Street Map initiative. We anticipate that it would be a valuable addition to the research community for the development of sophisticated approaches in the field of population estimation.
\end{abstract}


\section{Background \& Summary}  
Rapid urbanization in cities is leading to environmental concerns such as climatic changes, food and water scarcity, poor air quality, deforestation, etc \cite{mcdonald2011urban, tatem2014mapping, mcgranahan2007rising, zhang2022linking, szabo2016urbanisation}. To understand the key trends in urbanization, population estimation plays a crucial role.
 Traditionally, population estimation is done through a census. In this procedure, the population data is systematically collected and compiled over an administrative unit or a census unit. The accuracy of the data depends upon the number and size of the administrative unit, the collection method, the completeness of the survey, and varies significantly within the rural and urban regions \cite{leyk2019spatial}. 
 
 In recent years, statistical and machine learning methods have been applied directly to remote sensing data to estimate the population distribution \cite{wu2005population, leyk2019spatial}. In general, these methods have been applied either to a smaller region or based on some country-specific data such as building footprints or other detailed geospatial data sets which are not easily available in other countries. Stevens et al. \cite{stevens2015disaggregating} used a random forest approach to estimate the population at $\sim$ 100\,m resolution for Vietnam, Cambodia, and Kenya. They have incorporated a wide range of remotely-sensed and geospatial datasets, such as distance to roads, health facility, elevation, land cover, vegetation, settlements, and nighttime lights, and used the country-specific census data collected from the National Institute of Statistics for Cambodia, the National Statistics Office in Vietnam, and the National Bureaus of Statistics for Kenya. Doupe et al. \cite{doupe2016equitable} proposed a new method that used a Convolutional Neural Network (CNN) to estimate the population by combining Landsat-7 satellite imagery with (DMSP/OLS) nighttime lights. They trained their model on data from Tanzania at a 250\,m satellite pixel resolution and estimated the population for Kenya at an 8\,km resolution. They have published the code to reconstruct their data set for Tanzania and Kenya. Another similar CNN approach has been proposed by Robinson et al. \cite{robinson2017deep} They prepared their data from US census summary grids in combination with Landsat imagery to estimate the population in the US counties at a 1\,km resolution. Hu et al. \cite{hu2019mapping} also suggested a deep learning approach by combining satellite imagery from Landsat-8 and Sentinel-1 and used the Socio-Economic Caste Census survey to derive the population density for India. In most of the methods above, either the data is not available for download or could be reconstructed only for a few cities. Other gridded population products include Global Human Settlement Population Grid (GHS-POP) \cite{freire2016development}, WorldPop \cite{worldpop2018school}, Oak Ridge National Laboratory’s LandScan \cite{bhaduri2002landscan}, and High-Resolution Settlement Layer (HRSL) \cite{layerfacebook}, and so on. The difference in the estimation method and ancillary data used in the data sets lead to different results \cite{leyk2019spatial}. Few studies have been conducted to assess and compare the accuracy of gridded population products by comparing their estimations with the actual population counts \cite{chen2020multiple, sliuzas2017assessing}. However, these studies require collecting and processing the census data. Therefore, it becomes difficult and time-consuming to reproduce the results or compare the methods.
 
With our data set, we aim to fill these gaps by providing a systematic regression and classification scheme for population estimation in 98 European cities. The cities cover 28 European Union (EU) member states and the four EFTA countries. It represents a wide range of topography, demography, and architectural designs across the countries. It would save the cost of collecting and processing of a new data set to develop and validate the methods. The data set comprises digital elevation models (DEM), local climate zone (LCZ), land use (LU), and nighttime lights (VIIRS) in combination with multi-spectral Sentinel-2 imagery (SEN2), and data from the Open Street Map initiative (OSM). This multi-data source combination has not been explored before in the domain of population estimation. We expect that it will be a valuable addition to the research community for developing sophisticated approaches in the field of population estimation.
 
In this paper, we contribute to the current literature by providing a benchmark data set created from publicly available data sets. We investigated the fusion of multi-source data over a large number of cities. To demonstrate the potential capability of our data set, we trained the Random Forest model using the extracted features from the input data to estimate the population on our test data set. The initial results indicate that there is a conceivable potential for the development of powerful machine learning methods with the So2Sat POP data set.

\section{Methods}
\begin{figure}[ht]
    \centering
    \includegraphics[scale=0.65]{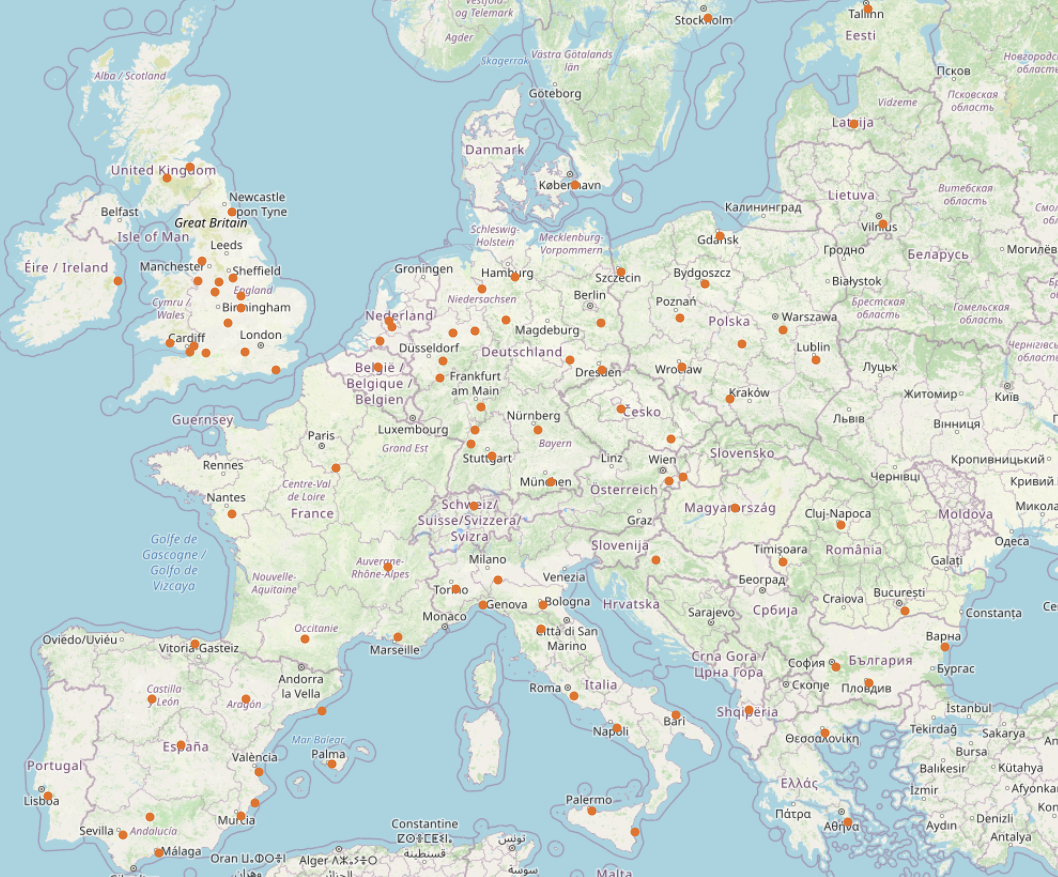}
    \caption{The orange dots on the figure above indicate the location of selected EU cities in our study.}     
    \label{fig:selected_europe_cities}
\end{figure}

Our region of interest (ROI) is spread over Europe (Figure~\ref{fig:selected_europe_cities}). Initially, we select all the cities across Europe with 300,000 inhabitants or more in 2014 as per the UN World Urbanization Prospects - The 2014 Revision \cite{united_nations_world_2014}. Out of these cities, we select 106 cities based on the availability of population data. Typically, a city can be described as a permanent large human settlement defined by the administrative boundaries. However, defining an administrative boundary could be very tricky because it changes as the census tracts merge or split over time. The outward expansion of the cities is far beyond their formal administrative boundaries \cite{habitat2013state, taubenbock2019new}. So, we employed an algorithm to determine the extent of the city. The city center coordinates listed in the UN World Urbanization Prospects - The 2014 Revision \cite{united_nations_world_2014} have been used as a starting point in combination with Global Urban Footprint (GUF) \cite{esch2017breaking} which provides a binary mask of urban vs. non-urban regions. A rectangle centered at the coordinate of each city is adaptively grown outwards until half of the area of the rectangle is not built up anymore according to the GUF. To take the rapid urbanization into account, each side of the rectangles is expanded by a factor of two (i.e. a factor of four in the area). Since the resulting rectangles of two neighboring cities might intersect with another, a set of rules is employed to allocate the intersecting area to one of the two cities and ensure that each city's extent covers a unique area. This set of rules is summarized in Algorithm \ref{alg:allocation}. The algorithm runs recursively and in descending order, starting from the cities with the highest relative overlap. According to the set of rules, $city_{a}$ is either merged into $city_{b}$ or the overlapping are is allocated to $city_{a}$ and removed from $city_{b}$, depending on the relative size of the overlapping area. The number of cities is merged to 98 in order to handle these overlaps. As per the defined extent of each city, we processed and prepared the following input data sources.

\begin{algorithm}
    \caption{Allocation of intersecting areas - Pseudocode.} \label{alg:allocation}
    \begin{algorithmic}[1]
        \Require $city_{b} \geq city_{a}$
        \For {$city_{a} \cap city_{b}$ in data set}
            \While {Highest overlap > $0.5*city_{a}$}
                \If {$city_{a} \subset city_{b}$}
                   \State Remove $city_{a}$
                \Else
                    \State Merge $city_{a}$ into $city_{b}$
                \EndIf
            \EndWhile
            \State Remove overlapping area from $city_{b}$ 
        \EndFor
    \end{algorithmic}
\end{algorithm}

\subsection{Input data sources}

\subsubsection{Population data}
Generally, the country-specific Census Bureau provides the population data by the hierarchy of administrative units. The heterogeneity in the size of the administrative units makes it difficult to use the data directly for the analysis. In such a scenario, gridded population products with uniformly sized grid cells offer detailed and consistent population maps. The most common way to produce the gridded population products is to redistribute the census count from the administrative level to the fine grid cells, conditioned by ancillary data such as land use and land cover maps, this method is also known as dasymetric mapping \cite{langford2007rapid}. 

The European Statistical System (ESSnet) project, in cooperation with the European Forum for Geography and Statistics (EFGS), aimed to produce the high resolution (1\,km) population grids from the population census in Europe. The methodology constitutes aggregation, disaggregation, and hybrid approaches based on the availability of the data. Aggregation (bottom-up approach) is assumed to be the best way to produce the population grids \cite{gallego2010population}. In this project, approximately 18 countries are using the aggregation or at least a hybrid method to produce the grid statistics \cite{efgs_final2011}. Due to a lack of geocoded microdata, the disaggregation (top-down approach) method has been employed in some regions. In disaggregation, the difference between the modeled population and the actual population of the region depends upon the size of the administrative unit. Often higher misplacement of the people is observed in the larger administrative unit. The overall quality of the product varies depending upon the data availability ranging from positional accuracy of 0.1m for each address and building in Austria to up to 100m in Estonia \cite{eurostat2011} (for details see the document named `GEOSTAT\_grid\_POP\_1K\_2011\_V2\_0\_QA.pdf' in the folder titled `GEOSTAT-grid-POP-1K-2011-V2-0-1'). The European Environment Agency published the population grids for EU28+ EFTA countries that comprise approximately 4.3 million km$^{2}$ with 480 million inhabitants \cite{gallego2010population}. It is freely available via Eurostat, for non-commercial purposes. Further details regarding the product standards, methodology, and quality assessments can be found at the GEOSTAT 1B project website \cite{efgs_quality2011}.  

\subsubsection{Sentinel-2}
The Sentinel-2 mission \cite{drusch2012sentinel} was launched in June 2015 by the European Space Agency (ESA) and consists of two identical satellites, 2A and 2B, phased at 180 degrees toward each other. Sentinel-2 satellites provide multi-spectral optical images with a span of 13 spectral bands at a spatial resolution of 10\,m, 20\,m, 60\,m. Thus, freely available Sentinel-2 imagery offers great potential for the fine-scale mapping of human settlements. The analysis-ready cloud-free mosaics of Sentinel-2 data have been achieved using the cloud detection approach based on a pixel-wise analysis \cite{Aggregating}. We have used all four seasonal sets of Sentinel-2 images to capture the seasonal variation in the data. The spring, summer, and autumn seasons are from 2017 and winter is from 2016.

\subsubsection{TanDEM-X Digital Elevation Model}
The goal of the TanDEM-X mission is to create a high-quality 3D topographic map of the Earth that is homogeneous in quality and unparalleled in accuracy. The data acquisition period for the global DEM product is between December 2010 and January 2015 and the global DEM has been produced in September 2016. With the coverage of 150 Million km$^2$ of the complete Earth’s landmasses and 10 m absolute height accuracy (90\% linear error) \cite{wessel2018accuracy}, it is suitable for various applications in Environmental research such as land cover and land use analysis, urban planning, climate change, etc. Currently, the TanDEM-X digital elevation model is the most accurate digital elevation model (DEM) available globally \cite{esch2020towards}. We have used the freely available TanDEM-X 90\,m (3 arcsec) DEM global product \cite{tandem90m} that contains the final, global Digital Elevation Model (DEM) of the landmasses of the Earth \cite{wessel2018tandem}.
        
\subsubsection{Local climate zones}
Local climate zones (LCZ) are a freely available systematic housing density classification data, formally developed to standardize urban heat island studies \cite{stewart2011local}. Based on the land surface and properties, it consists of 17 structural types of which 10 are describing built zones from compact high-rise to open low-rise and 7 are natural zones ranging from dense vegetation to bare lands. Thus, each zone is characterized by the built-up and land cover properties. We have used the urban local climate zone classifications, So2SatLCZ v1.0 at the resolution of 100\,m, produced by fusing the freely available satellite data from Sentinel-1 and Sentinel-2 satellites using deep learning \cite{zhu2019so2sat}. The patches in this data set are hand-labeled as per the local climate zones classification scheme by 15 domain experts, followed by a visual and quantitative evaluation process over six months. This benchmark data set, therefore, is of potential use for urbanologists, demographers, climatologists, and many other researchers.

\subsubsection{Nighttime lights}
Observational Nighttime lights (NTL) show a strong correlation with the spatial distribution of the human population \cite{liu2011relationships}. We studied the two widely used Nighttime lights data, the DMPS-OLS and the NPP-VIIRS satellite images. NPP-VIIRS with a better spatial resolution (15 arc-second, about 500\,m) has a higher potential in modeling the socio-economic indicators \cite{shi2014evaluating}. The annual composites of VIIRS nighttime
lights derived from monthly mean data are freely available from 2012 to
2019 (\url{https://eogdata.mines.edu/products/vnl/}). We used a masked average radiance version of VNL V2, which is a preprocessed version free of outliers from fleeting events \cite{elvidge2021annual}. 

\subsubsection{OSM}
OpenStreetMap (OSM) (\url{http://www.openstreetmap.org}) is an open and crowd-sourced platform for maps, available under the Open Data Commons Open Database License (ODbL) (\url{http://www.opendatacommons.org/licenses/odbl/1.0/}). Geo-referenced locations on a very detailed level can be entered as nodes, ways, or relations and specified with informative tags. These locations include any type of buildings, streets, boundaries, water bodies, etc.\cite{osmwiki}. Our data set contains low-level and high-level features derived from OSM data. At a low level, a locale with a high number of certain nodes (e.g. supermarkets, gas stations, residential buildings, schools, etc.) correlates strongly with the population in the vicinity. A simple counter statistic of such node-types is a strong indicator and can serve as a feature vector for population density estimation.
As high-level features we extracted the building functions from OSM building tags to represent urban land use. This information is closely linked to employment, social support, and population \cite{li2016physical} and indicates the interaction between human activities and the environment.

\subsection{Data Preprocessing}
\begin{figure}[ht]
    \centering
    \includegraphics[scale=0.48]{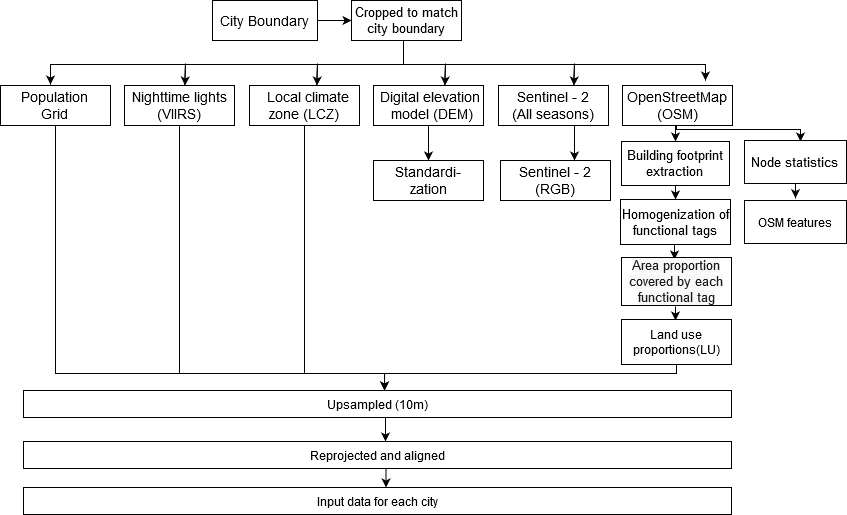}
    \caption{Step-by-step preprocessing of all the input data sources to prepare the corresponding input data for each city.}     
    \label{fig:ancillary_data_flowchart}
\end{figure}

We employed two-step preprocessing for all input data sources. In the first step, input data for each city has been created. The second step includes the creation of the 1x1\,km patches for each city. Figure~\ref{fig:ancillary_data_flowchart} outlines the step-by-step preprocessing of all the data that has been employed to create the input data for each city. All the input data has been cropped using the city boundaries defined by our algorithm. DEM data is standardized by subtracting the DEM mean and then scaled to unit variance. To match the spatial resolution of the Sentinel-2 RGB bands, other data sources such as the digital elevation model (DEM), local climate zone (LCZ), land use (LU), and VIIRS nighttime lights (VIIRS) have been up-sampled to 10\,m. Since the input data sets are from various sources, they are in different Coordinate Reference Systems (CRS). The VIIRS is in WGS84 (EPSG:4326), the LCZ, DEM, and SEN2 data are in the Universal Transverse Mercator (UTM) zone and the population grid is in the EPSG:3035 - ETRS89-extended / LAEA Europe. To align the input data with the population grid, all the input data have been reprojected from their corresponding coordinate system to the EPSG:3035 coordinate reference system.

\subsubsection{Low-Level Features}
  The OSM planet dump of 2017-07-03 is being downloaded directly from the OpenStreetMap archive (\url{https://planet.osm.org/planet/2017/}). To reduce the computation time by touching the huge planet file only once, we extract the bounding box of each corresponding city and subsequently 1x1\,km patches from these extracted city-dumps with the \texttt{Osmosis command line tool}(\url{https://github.com/openstreetmap/osmosis}). The OSM node statistics are extracted for each 1x1\,km patch of the city using the OSMnx python Library \cite{BOEING2017126}. Table 1 shows the considered OSM tags over which the statistical analysis has been done.
 
\begin{table}[htbp]
\centering
\begin{tabular*}{\textwidth}{@{\extracolsep{\fill}}cccccccc@{}}
\toprule
aerialway & boundary  & geological & landuse   & natural & public\_transport & sport   & waterway         \\ 
aeroway   & building  & healthcare & leisure   & office  & railway           & telecom & addr:housenumber \\
amenity   & craft     & highway    & man\_made & place   & route             & tourism & restrictions     \\
barrier   & emergency & historic   & military  & power   & shop              & water   & other: True      \\ \bottomrule
\end{tabular*}
\caption{Nodes with these OSM tags are considered for the statistical analysis/counting of the corresponding 1x1\,km patch.}
\label{tab:osm-tags}
\end{table}

\subsubsection{High-Level Features}
\begin{figure}[ht]
    \centering
    \includegraphics[width=\textwidth]{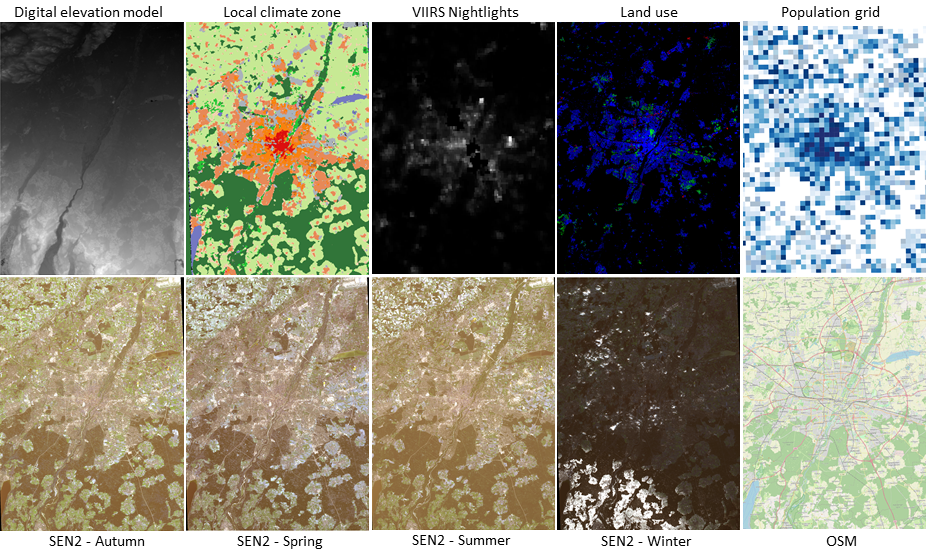}
    \caption{All the input data for the  Munich city, created using the first step of data preprocessing.}     
    \label{fig:ancillary_rasters_example}
\end{figure}

 To create the land use data we analyzed three different tags of buildings in OSM: \textit{building}, \textit{amenity}, and \textit{shop}. For each of them, OSM provides a guideline on possible values. In total, there are 341 possible values for these three tags, which are mapped to a homogenized and simplified land use classification scheme: \textit{commercial}, \textit{industrial}, \textit{residential}, and \textit{other}. As the three tags can occur jointly, we make sure that they are not contradicting each other and omit buildings that have inconsistent values. Moreover, the tags are captured as free-form text fields and hence, OSM contributors are not restricted to use them but can enter any text. After homogenizing the semantic information, we convert the vector data of building polygons into raster data. The rasterized value represents the area covered by building polygons inside a raster pixel. For scaling, we divided the area by the area of a pixel to yield a relative number of how much of a pixel is covered by building polygons. Applying this procedure for each land use class results in a four-band raster with corresponding land use proportions. The output of the first step of data preprocessing for Munich city can be seen in Figure~\ref{fig:ancillary_rasters_example}.

 In the second step, for each city, we created the patches using all the input data processed in the first step. The population grid of a city is used as a reference grid to crop all the other input data. The size of a grid cell in the population grid is 1x1\,km and each cell represents the population count living per square km of the cell. The grid cells along the border of the population grid might belong to two adjacent cities. To avoid this duplicity such that one grid cell should belong to only one city, we applied an area thresholding to all the cells. Only the cells with an area greater than 900,000\,m$^{2}$ or size $\sim$ 0.95 x 0.95\,km have been considered as a part of the city. This eliminates the grid cells from the edges which are not fully included in the city boundary. Also, the reference Geostat population grid has some missing cells. The missing cells mostly contain the uninhabited regions. Such regions cover the green fields and water bodies. We included these missing cells in our data set to allow for zero population predictions. The patches cropped using the population grid have been assigned the population count as of the corresponding grid cell. The rest of the missing\slash uninhabited patches have been assigned as zero population count. For some applications such as environmental impact assessments, land use analysis, climate change, etc. it is sufficient to know the range of the people living in a region. So, we further preprocessed the population grids by binning the population count to a population class. We assigned a grid cell, Class 0, if the population count of the cell is zero, C$_{cell}$=0 if P$_{cell}$=0 and subsequently C$_{cell}$=1 if 2$^{0}$ $\leq$ P$_{cell}$ ${<}$ 2$^{1}$, C$_{cell}$=2 if 2$^{1}$ $\leq$ P$_{cell}$ ${<}$ 2$^{2}$, C$_{cell}$=3 if 2$^{2}$ $\leq$ P$_{cell}$ ${<}$ 2$^{3}$.....C$_{cell}$=${k+1}$ if 2$^{k}$ $\leq$ P$_{cell}$ ${<}$ 2$^{k+1}$ where k $\in$ $\mathbb{N}$. For our data, the highest value of ${k}$ is 16. This process of discretization is inspired by Robinson et al. \cite{robinson2017deep}. Thus, each grid cell has been assigned a population class in addition to the absolute population count. It would give more flexibility to the end-users to develop either regression or a classification model for the task considering the requirements of the application.

\begin{figure}[ht]
    \centering
    \includegraphics[scale=0.60]{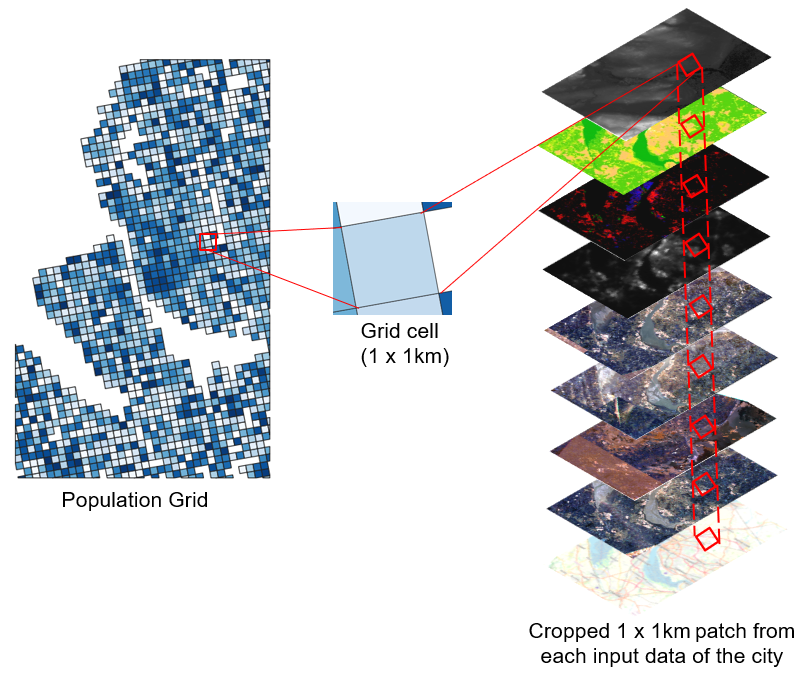}
    \caption{Patch creation process, second step of data preprocessing. All input data sources have been cropped for each cell in the population grid. The size of each patch is 1 x 1\,km.}     
    \label{fig:patch_creation}
\end{figure}

\begin{figure}[ht]
    \centering
    \includegraphics[scale=0.65]{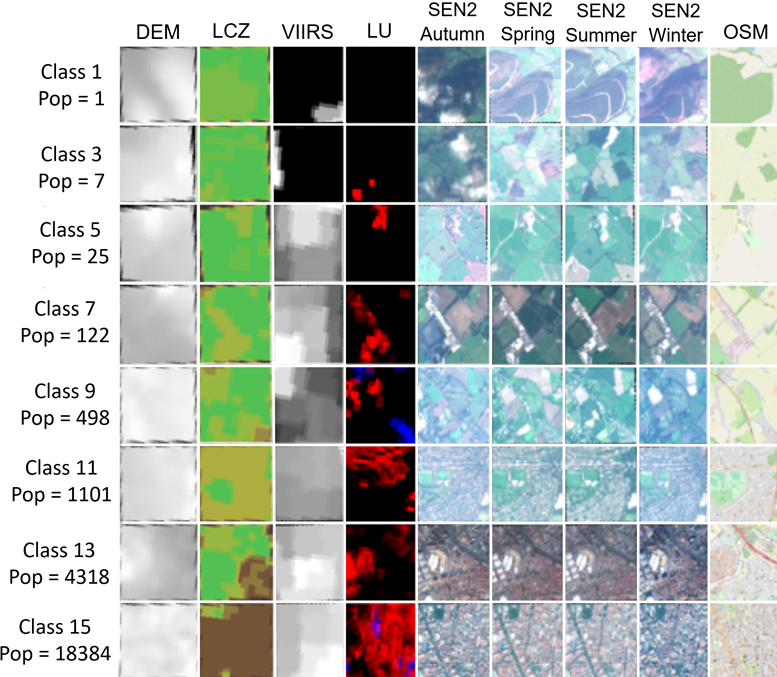}
    \caption{Sample patches from the odd numbered classes of our data set. Lower classes depicts sparsely populated regions while higher classes depicts densely populated regions.}     
    \label{fig:Upto_class13_patch_example}
\end{figure}

Figure~\ref{fig:patch_creation} illustrates the patch creation process. For every grid cell, a total of 9 patches, one from each data source have been created. We called the 9 patches corresponding to one grid cell, a patch-set. Each patch-set represents a population count as of the corresponding population grid cell and a population class depending on which bin the population count of the grid cell falls at the resolution of 1\,km. Figure~\ref{fig:Upto_class13_patch_example} depicts the odd-numbered class samples from our data set with their corresponding patch-set, population class and population count. The lower classes represent sparsely populated regions. Patches belonging to lower classes mostly contain green fields, water bodies, and bare land. As the class number goes higher, patches contain few low-rise to dense high-rise built-up regions. In other words, patches from lower to higher classes represent rural to urban regions.

\section{Data Records}
The final data set consists of two parts, So2Sat POP Part1 \cite{So2Sat_POP_Part_1} and So2Sat POP Part2 \cite{So2Sat_POP_Part_2}. All the data patches except OSM data are available as GeoTiff images. Along with the raw osm patches, we also provide the features extracted from the OSM data as separate Comma Separated Value (CSV) files. So2Sat POP Part1 consists of the patches from local climate zones, land use, nighttime lights, Open Street Map features, and from all seasons(autumn, summer, spring, winter) of Sentinel-2 imagery, a total of 1,104,688 patches. So2Sat POP Part2 consists of patches from the digital elevation model and Open Street Map only, a total of 276,172 patches. So2Sat POP Part1 has the storage requirement of $\sim$ 98 GB and So2Sat POP Part2 requires $\sim$ 5.20 GB.

\subsection{Data set structure and naming convention}
Both parts of the data set consist of a predefined train and test split. Out of 98 cities, 80 cities ($\sim$ 80\% of the data) have been randomly selected as a training set and the rest of the 18 cities ($\sim$ 20\% of the data) constitucorrecte the test set. The city folder has been named as \texttt{xxxx\_xxxxx}\_city\_name, where \texttt{xxxx\_xxxxx} constitutes randomly generated identification number and the postal code of the city.

The digital elevation model (\texttt{DEM}), local climate zone (\texttt{LCZ}), land use (\texttt{LU}), nighttime lights (\texttt{VIIRS}), Sentinel-2 autumn data (\texttt{sen2autumn}), Sentinel-2 spring data (\texttt{sen2spring}), Sentinel-2 summer data (\texttt{sen2summer}), Sentinel-2 winter data (\texttt{sen2winter}), Open Street Map (\texttt{OSM}) and its corresponding extracted features (\texttt{osm\_features}) are the utilized input sources. All city folders in So2Sat POP Part1 contain 7 sub-folders, one for each input data source except the OSM and DEM data, a separate folder for processed osm features, and a comma-separated value (\texttt{*.csv}) file that contains the absolute population count and population class for each patch. In So2Sat POP Part2, the city folders contain the Open Street Map and digital elevation model data sub-folder. All data folders have their class sub-folders. The class folders have been named as \texttt{Class\_x} where \texttt{x} denotes the class number. The number of class folders in a city depends on its population distribution. For example, Malaga has the highest class folder as 16 because the highest population count in the 1x1\,km area of the city is 39535 while in Riga the highest population count is 15839 so the highest class folder in its city folder is 14. A patch has been assigned a unique identification code using the naming convention of its corresponding population grid cell. The naming of the grid cell (based on the LAEA grid) begins with the size of the cell (1\,km) followed by the coordinates (in km) of the lower left-hand corner, starting with the letter \texttt{N} followed by the latitude and \texttt{E} followed by the longitude, e.g.~\texttt{1kmN4101E4453} \cite{efgs_final2011}. The patches that do not correspond to the grid cells in the population grid have been simply given a numeric identification number.

\subsection{Access and availability}
The data set is available for download at the persistent links provided by the official media library of the Technical University of Munich (TUM). So2Sat POP Part 1 (\url{https://mediatum.ub.tum.de/1633792}) is distributed under the Creative Commons Attribution 4.0 International License (\url{http://creativecommons.org/licenses/by/4.0/}) and So2Sat POP Part 2 (\url{https://mediatum.ub.tum.de/1633795}) is distributed under the Creative Commons Attribution Share-Alike International License (\url{http://creativecommons.org/licenses/ by-sa/4.0/}). Kindly cite this paper when the data set is used.

\section{Technical Validation}
To demonstrate the suitability of the data set for population estimation, we implemented the popular Random Forest (RF) algorithm \cite{breiman2001random}, because of its flexibility, efficiency in handling the noisy input data, and relative resistance to overfitting \cite{grippa2019improving}. Another advantage of the random forest algorithm is that it is easy to measure the relative importance of each feature on the prediction. We implemented the supervised random forest algorithm in Python with the scikit-learn library for both the regression and classification tasks. We used grid search to automatically fine-tune the number of trees to grow and the maximum number of features to consider splitting a node and assess the performance through a 10-fold cross-validation. To train the model, different features are constructed from all the input data patches of 80 train-set cities. The constructed features include min, max, mean, median, and standard deviation from only the RGB bands of Sentinel-2 imagery, mean and max for digital elevation model and nightlights, the total area covered by each class in land use, majority class for local climate zone, and osm-based features such as street density, presence of highways, railways, etc., extracted from osm patches. Using this process, we calculated 125 features for each patch. For regression, we used the absolute population count as the response variable while in classification, class labels are used as ground truth. 

\begin{figure}[!tbp]
    \centering
    \subfloat[]{{\includegraphics[width=8cm]{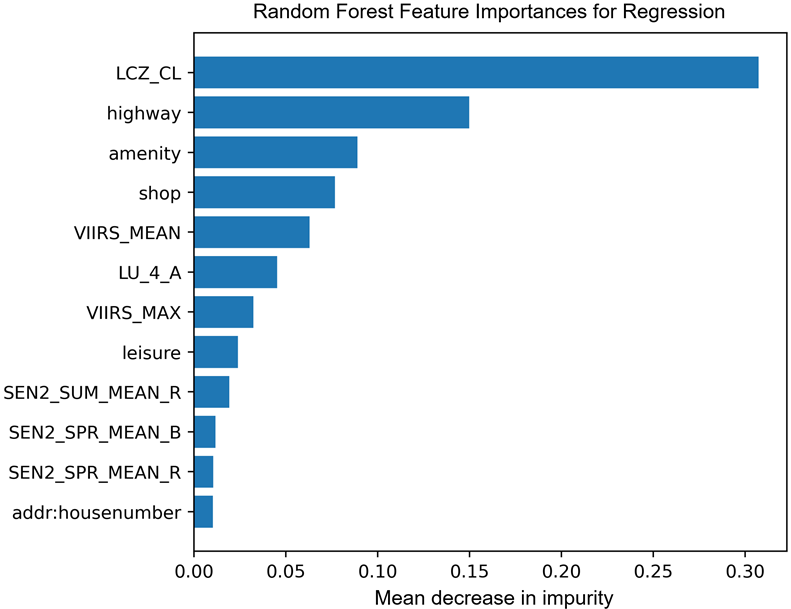} }}%
    \hfill
    \subfloat[]{{\includegraphics[width=8cm]{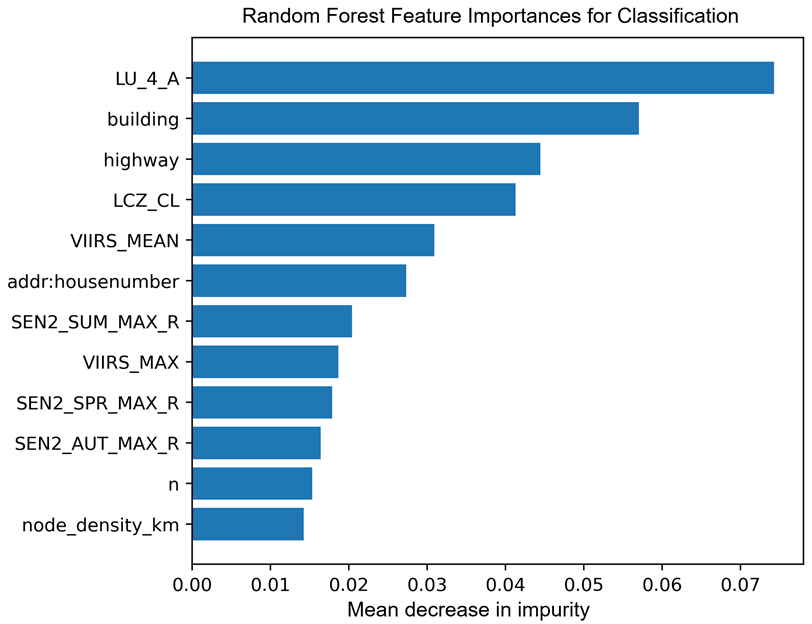} }}%
    \caption{Random Forest feature importances based on mean decrease in impurity (MDI). The higher the value the more important the feature. Plot shows only the twelve most relevant features for both regression (a) and classification (b)}%
    \label{fig:feature_importance}%
\end{figure}

\begin{figure}%
    \centering
    \subfloat[][\centering]{{\includegraphics[width=8.2cm]{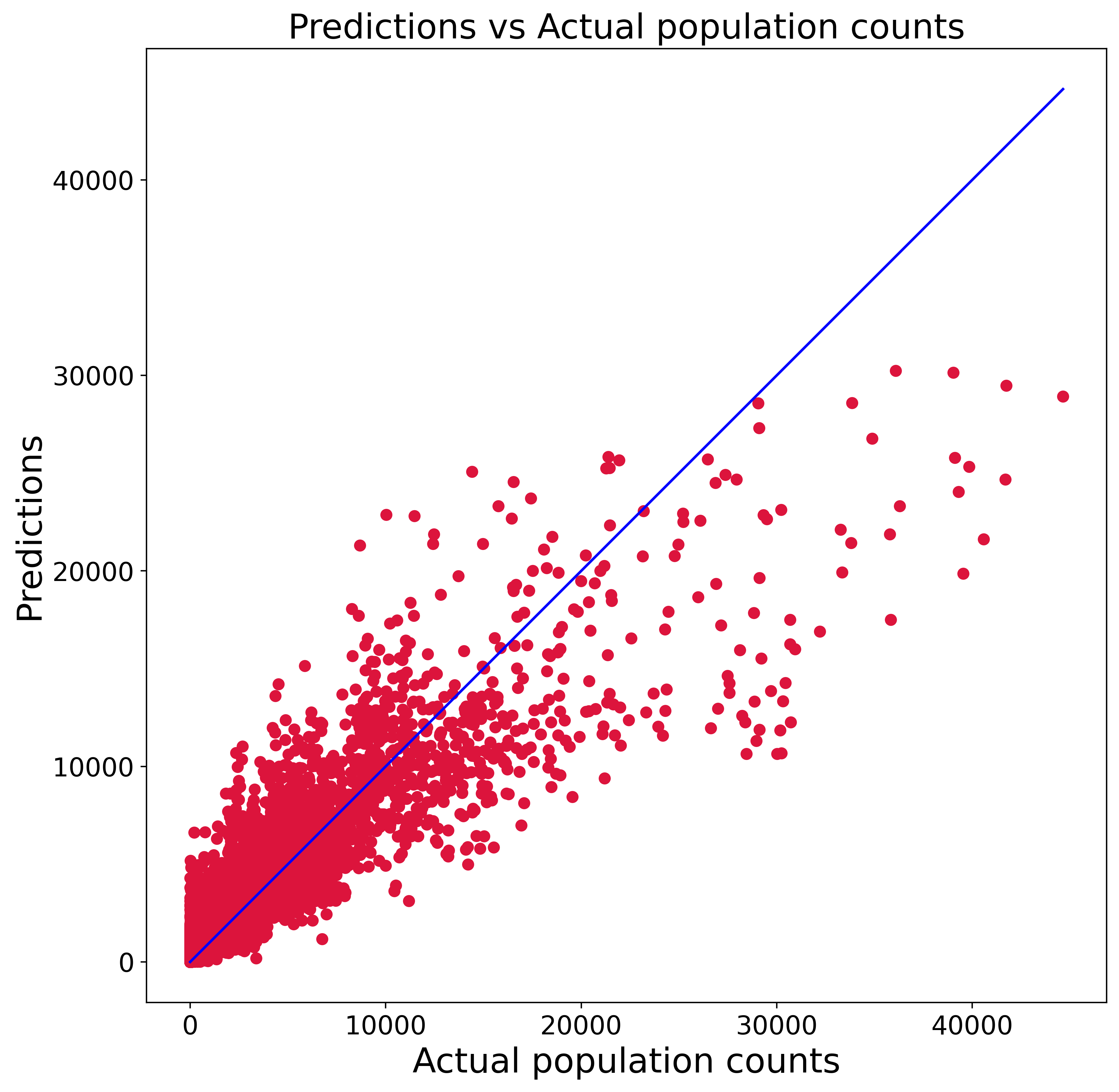} }}%
    \hfill
    \subfloat[][\centering]{{\includegraphics[width=8.8cm]{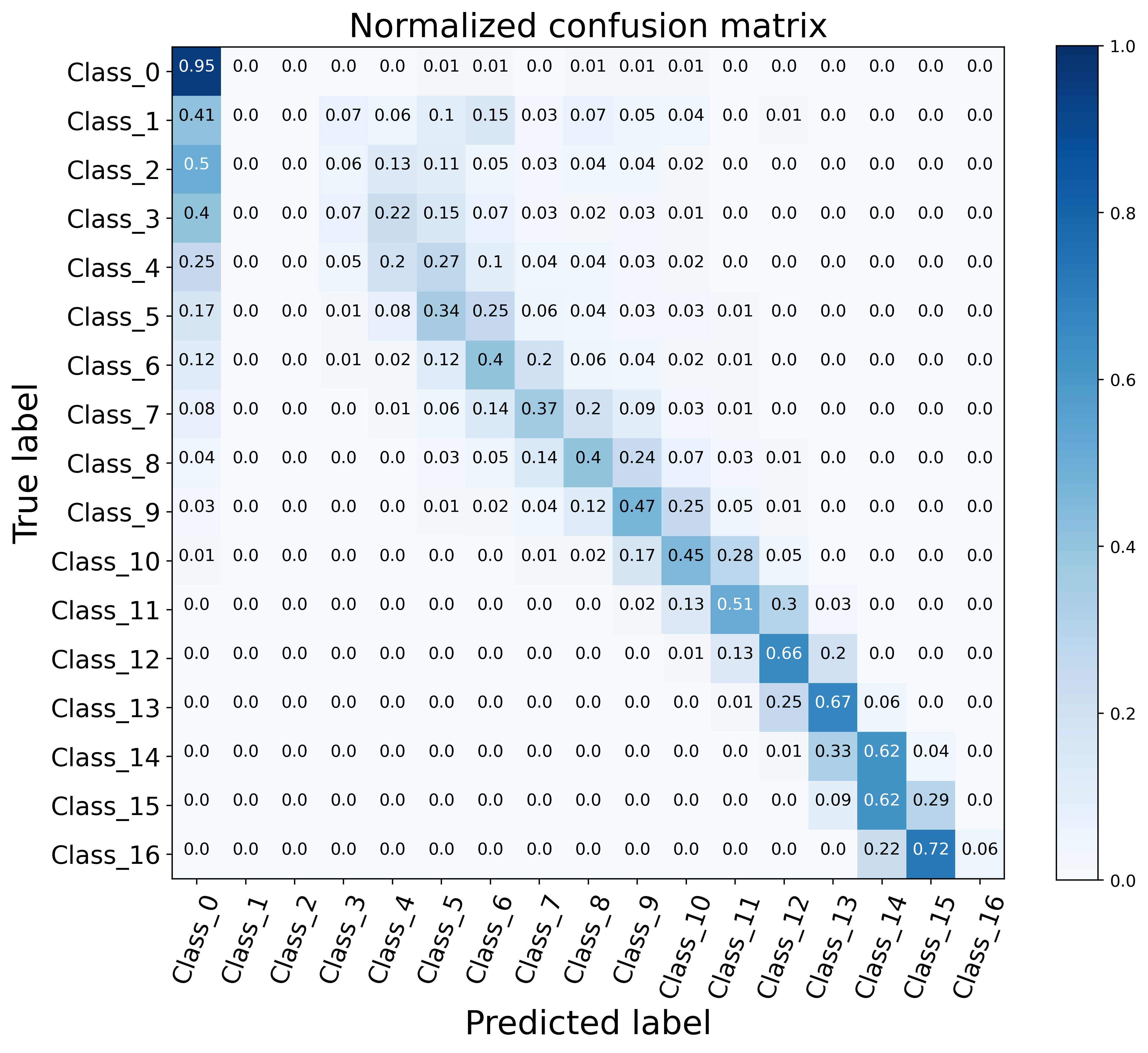} }}%
    \caption{(a) Predicted vs. Actual Values for regression, the model fits good except for the high population counts where the points appeared dispersed from regressed diagonal line (b) Confusion matrix for classification, normalized by class support size (number of patches in each class). Confusion among the non-urban classes is higher than among the urban classes.}%
    \label{fig:evaluations}%
\end{figure}

The trained model has been evaluated on the 18 unseen test cities. Figure~\ref{fig:feature_importance} shows only the twelve most relevant features that have been selected by the Random Forest algorithm and used to estimate the population count and a population class for the set of test data. The features extracted from the OSM data, nightlights, and LCZ classes have been ranked as the most important features for both tasks. For regression, we calculated the root-mean-square error (RMSE) and the mean absolute error (MAE). Table~\ref{table:2} indicates the performance of the regression model.
Since the data set is imbalanced due to the higher percentage of non-urban regions than the urban regions, we used balanced accuracy to evaluate the classification performance. Also, we used macro-averaged Precision, Recall, and F1-score metrics to treat each class equally, regardless of any imbalance. Table~\ref{table:1} summarized the results from the classification. To further describe the performance of the classification model, we plotted a normalized confusion matrix on the set of test data. Figure~\ref{fig:evaluations}\,(b) illustrates that the model is confident in predicting the higher classes (urban regions) while it does not perform well on the lower classes, which represent the sparsely populated regions. The initial three classes (Class 1, 2, and 3) represent the regions where the population count is in the range 1-8, and, likely, the features among these three classes are not distinguishable enough. So, it becomes very difficult to differentiate the patches from Class 1 to Class 3. One simple solution to it is the merging of these initial classes, which represent the population count range of less than 8. It could significantly improve the results without the loss of any critical information. To give more flexibility to the user, we provided the data set without any such post-processing, and it could easily be rearranged as per the specific needs. For regression, to visually assess the model fit, we plotted the actual population count of the grid cells versus the population count predicted for each grid cell of the test data. Figure~\ref{fig:evaluations} (a) indicates that the model underestimates the actual values for the patches where the population count is high while for the patches with a population count of less than 15.000, it is relatively a good fit. We believe that with more sophisticated features and machine learning methods, a powerful model could be developed to estimate the population using our data set.

\section{Usage Notes}
In this paper, we proposed a unique data set that combines multi-data sources which have not been explored at the cross-country level in this domain before. Our data set covers many cities across Europe thus offering diverse topography and demography. Also, the fusion of distinct data sources gives much more information about the landscape and other socio-demographic attributes of a region. 

The availability of precise and detailed population data varies from country to country. The population grid used as a reference grid in our data set is available throughout Europe at a consistent resolution of 1\,km, however, its precision may vary depending upon the quality of the data available in a country. Also, the reference population grid is based on the 2011 population and housing census data while other input data sources belong to a different time frame, for example, Sentinel-2 data belongs to the year 2016 as the mission itself has been launched in 2015. The time difference between the collection of the population data and other corresponding input data might introduce some noise to the evaluations. Usually, the population census has been conducted once in a decade, so it becomes very difficult to collect the other corresponding data from the same year, especially when collecting from multiple data sources. We still believe that this data set will be helpful in the development of the machine-learning based approaches in this domain. Also, the evaluation of different methodologies has always been a challenge due to the unavailability of the freely available benchmark data set. So, we hope this data set will be useful for the comparison and evaluations of the different approaches in the state of the art.

The data preparation is performed in Python using the common libraries for the geospatial data such as Geopandas, Fiona, Rasterio, and Geospatial  Data  Abstraction  Library (GDAL). Our computing system consists of an HPC Server with 2x AMD EPYC ROME 7402 CPUs (48 core) and needed 5-6 days to complete the OSM patch files and their statistics. Along with the patches, we also provided the Comma Separated Value (CSV) files for each city that contains the actual population count and its corresponding population class for each grid cell. Thus, the end user could rearrange the data set using its corresponding population count according to the requirements of the use case. 

We hope that this basic data set will make it possible to develop new statistical and machine learning approaches that derive the population in higher spatial resolution, but above in more consistency. It is precisely this consistency that is often lacking in today's population data sets across countries. This data set is intended to lay the foundation for improved comparative studies in various application domains. This improved database for population distribution may be central information both, in the academic and non-academic fields. Be it as necessary information for spatial or urban planning, such as the provision of living space or socio-technical infrastructure, be it in the locale of risk analysis or coordination in an event, be it in the validation, support, or updating of censuses, or be it in comparative studies on topics such as migration in general and, more specifically, on urbanization or sub-urbanization trends. 

\section{Code availability}
Python is used for all the analyses and implementations. The code to create the features for each city and to run the baseline experiments is available on our GitHub project (\url{https://github.com/zhu-xlab/So2Sat-POP}).

\section*{Acknowledgements} 

This research was funded by the European Research Council (ERC) under the European Unions Horizon 2020 research and innovation program with the grant number ERC-2016-StG-714087 (Acronym: So2Sat, project website: www.so2sat.eu), Helmholtz Association under the framework of the Helmholtz AI (grant  number:  ZT-I-PF-5-01) –Local Unit “Munich Unit @Aeronautics, Space and Transport (MASTr),” and Helmholtz Excellent Professorship “Data Science in Earth Observation – Big Data Fusion for Urban Research (grant number: W2-W3-100) and by the German Federal Ministry of Education and Research (BMBF) in the framework of the international future AI lab "AI4EO -- Artificial Intelligence for Earth Observation: Reasoning, Uncertainties, Ethics and Beyond" (Grant number: 01DD20001). Additionally, Sugandha Doda is supported by the Helmholtz Association under the joint research school “Munich School for Data Science - MUDS”. 

\section*{Author contributions statement}
S.D. drafted the manuscripts, S.D., E.H., M.K. and K.O. undertook data preparation, S.D. implemented and performed experiments, S.D. performed technical validation of the results. S.D., X.Z., Y.W., H.T., and M.K. edited the manuscript. All authors read and approved the final version of the manuscript.

\section*{Competing interests}
The authors declare no competing of interest.

\section*{Figures \& Tables}

\begin{table}[ht]
\begin{center}
\begin{tabular}{
 |p{2cm}||p{3cm}|p{1.5cm}|p{1.5cm}|  }
 \hline
 \multicolumn{4}{|c|}{Regression} \\
 \hline
Method& Internal OOB Score &RMSE &MAE \\
 \hline
 RF   & 0.864 &1276.26 &463.35 \\
 \hline
\end{tabular}
\caption{Evaluation of Random Forest model to estimate the population count on test data set.}
\label{table:2}
\end{center}
\end{table}

\begin{table}[ht]
\begin{center}
\begin{tabular}{
 |p{2cm}||p{1.5cm}|p{3cm}|p{1.5cm}|p{1.5cm}|p{1.5cm}|  }
 \hline
 \multicolumn{6}{|c|}{Classification} \\
 \hline
Method& Accuracy & Balanced Accuracy &F1 score &Precision &Recall\\
 \hline
 RF   & 0.5913 &0.3795 &0.3833  &0.4533 & 0.3795\\
 \hline
\end{tabular}
\caption{Evaluation of Random Forest model to predict the population class on test data set.}
\label{table:1}
\end{center}
\end{table}

\bibliographystyle{unsrt}  
\bibliography{template}  


\end{document}